\documentclass[letterpaper]{article} 

\usepackage{aaai24}  

\usepackage{times}  
\usepackage{helvet}  
\usepackage{courier}  
\usepackage[hyphens]{url}  
\usepackage{graphicx} 
\usepackage{natbib}  
\usepackage{caption} 
\usepackage{makecell}
\usepackage{algorithm}
\usepackage{algorithmic}

\usepackage{newfloat}
\usepackage{listings}
\DeclareCaptionStyle{ruled}{labelfont=normalfont,labelsep=colon,strut=off} 
\floatstyle{ruled}
\newfloat{listing}{tb}{lst}{}
\floatname{listing}{Listing}
\usepackage{amsmath}
\usepackage{multirow}
\usepackage{booktabs}
\usepackage{tcolorbox}

\usepackage{amssymb}
\NewDocumentCommand{\codeword}{v}{%
\texttt{\textcolor{blue}{#1}}%
}

\title{Reformulating Vision-Language Foundation Models and Datasets \\Towards Universal Multimodal Assistants}

\author{
    Tianyu Yu\equalcontrib\textsuperscript{\rm 1}  
    \hspace{2mm}Jinyi Hu\equalcontrib\textsuperscript{\rm 1}  
    \hspace{2mm}Yuan Yao\textsuperscript{\rm 1}\thanks{Corresponding authors.}  
    \hspace{2mm}Haoye Zhang\textsuperscript{\rm 1}  
    \hspace{2mm}Yue Zhao\textsuperscript{\rm 2}  \\
    Chongyi Wang\textsuperscript{\rm 3}  
    \hspace{2mm}Shan Wang\textsuperscript{\rm 3} 
    \hspace{2mm}Yinxu Pan\textsuperscript{\rm 4} 
    \hspace{2mm}Jiao Xue\textsuperscript{\rm 3} 
    \hspace{2mm}Dahai Li\textsuperscript{\rm 3} \\
    Zhiyuan Liu\textsuperscript{\rm 1}$^\dag$ 
    \hspace{2mm}Hai-Tao Zheng\textsuperscript{\rm 1}$^\dag$ 
    \hspace{2mm}Maosong Sun\textsuperscript{\rm 1}$^\dag$
}
\affiliations{
    \textsuperscript{\rm 1}Tsinghua University  \hspace{2mm} \textsuperscript{\rm 2} Beijing University of Posts and Telecommunications\\ \textsuperscript{\rm 3} Zhihu Inc.\hspace{2mm}  \textsuperscript{\rm 4} ModelBest Inc.

    {\tt yiranytianyu@gmail.com}
}

\usepackage{bibentry}

\begin{document}

\maketitle

\begin{abstract}

Recent Multimodal Large Language Models (MLLMs) exhibit impressive abilities to perceive images and follow open-ended instructions. 
The capabilities of MLLMs depend on two crucial factors: the model architecture to facilitate the feature alignment of visual modules and large language models; the multimodal instruction tuning datasets for human instruction following.
(\romannumeral1) For the \textit{model architecture}, most existing models introduce an external bridge module to connect vision encoders with language models, which needs an additional feature-alignment pre-training. In this work, we discover that compact pre-trained vision language models can inherently serve as ``out-of-the-box'' bridges between vision and language. Based on this, we propose Muffin framework, which directly employs pre-trained vision-language models to act as providers of visual signals.
(\romannumeral2) For the \textit{multimodal instruction tuning datasets}, existing methods omit the complementary relationship between different datasets and simply mix datasets from different tasks. Instead, we propose UniMM-Chat dataset which explores the complementarities of datasets to generate 1.1M high-quality and diverse multimodal instructions. We merge information describing the same image from diverse datasets and transforms it into more knowledge-intensive conversation data. 
Experimental results demonstrate the effectiveness of the Muffin framework and UniMM-Chat dataset. Muffin achieves state-of-the-art performance on a wide range of vision-language tasks, significantly surpassing state-of-the-art models like LLaVA and InstructBLIP. Our model and dataset are all accessible at https://github.com/thunlp/muffin.
\end{abstract}

\section{Introduction}







\begin{figure}[t]
    \centering    \includegraphics[width=0.9\columnwidth]{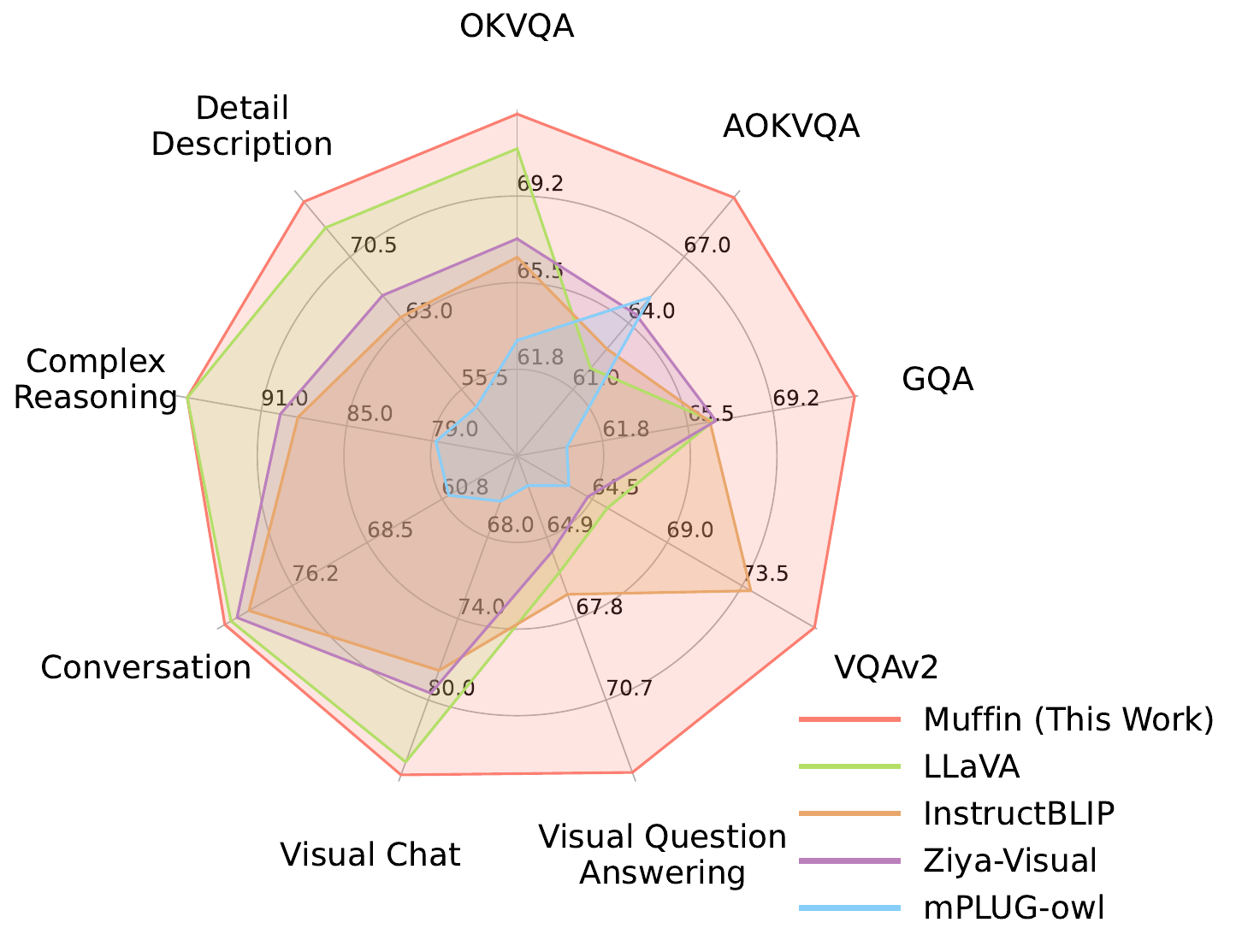}
    \caption{Muffin achieves state-of-the-art performances on various tasks compared with strong MLLMs. Visual Question Answering: the average score over four visual question answering datasets. Visual Chat: the average score over the conversation task, the complex reasoning task and the detail description generation task.}
    \label{fig:radar}
\end{figure}

Building a general model capable of tackling diverse tasks across multiple modalities has remained a longstanding goal within the realm of Artificial Intelligence. Recently, powerful Multimodal Large Language Models (MLLMs) have emerged as one of the most promising ways to achieve this goal, such as MiniGPT-4 \citep{zhu2023minigpt}, LLaVA \citep{liu2023visual}, and InstructBLIP \citep{instructblip}. These models empower large language models (LLMs) with impressive multimodal instruction-following capabilities by equipping LLMs with vision encoders to perceive visual content. 

Despite existing capabilities, several crucial factors of developing MLLM are still under-explored. In this work, we focus on two key challenges of building MLLMs: (\romannumeral1) effectiveness of model architectures to achieve feature alignment; (\romannumeral2) construction of multimodal instruction tuning dataset. 

For the model architecture, existing MLLMs can roughly be summarized as two streams: (1) A linear projector is optimized to align the frozen visual encoder with the frozen LLM, such as LLaVA \citep{liu2023visual} and PaLM-E \citep{driess2023palm}; (2) A visual feature re-sampler \citep{alayrac2022flamingo, li2023blip, instructblip} is optimized to compress the output of the visual encoder into a fixed-length feature sequence and align the these features with LLMs.
However, merely using a linear projector restrains the model's capacity to learn new knowledge and the feature sequence length is quadratically related to the resolution of the input image, leading to a significant computational burden. On the other side, introducing a visual feature re-sampler requires a resource-consuming additional training process to primarily achieve the alignment of modalities before connecting the visual encoder with LLMs \citep{li2023blip}.





To address the aforementioned limitations, we propose Muffin\footnote{\underline{Mu}ltimodal \underline{f}oundation models are \underline{f}ound to be ``out-of-the-box'' multimodal \underline{in}terfaces for LLMs.}, an efficient architecture to build powerful MLLMs. Intuitively, we notice that compact pre-trained vision-language models (VLMs), such as ALBEF \citep{li2021align}, CoCa \citep{yu2022coca}, and BEiT-3 \citep{wang2023image}, have already exhibited remarkable performance in vision-language (V-L) tasks through pre-training on extensive multimodal datasets. As a result, these VLMs inherently achieve the alignment of modalities and are potentially competent as ``out-of-the-box'' bridge modules to empower LLMs with visual capabilities. Based on this intuition, Muffin directly leverages pre-trained VLMs and learns a set of query vectors in the embedding space of VLMs to perceive the visual representation for LLMs. In this way, we are able to directly optimize the visual module to connect with LLMs without losing capacity or undergoing the additional alignment process. Experimental results show that Muffin can achieve the state-of-the-art performance among existing MLLMs.





In terms of the construction of multimodal instruction tuning datasets, most recent works \citep{gong2023multimodal, instructblip} simply formulate the downstream vision-language dataset into a unified format, while the short and limited format of responses in these datasets will harm the generative abilities of LLMs. Another line of works \citep{zhu2023minigpt, liu2023visual} converts isolated datasets into conversation corpora based on ChatGPT or GPT-4 \citep{openai2023gpt4}. However, they neglect the compementarities of different datasets which is crucial to form a comprehensive view of the image content and consequently lead to knowledge scarcity in generated data. 

To overcome such shortcomings, we design a simple and effective approach to reformulate multiple datasets into chat corpora with a flexible format in responses. Therefore, despite the lack of information in one annotation, multiple annotations for the same image can be complementarily merged to form a more comprehensive description of the image. Specifically, we use images from COCO \citep{lin2014microsoft} to construct the dataset. Based on the combined annotations, we require ChatGPT to generate high-quality chat corpora that are accurate and knowledge-intensive. Following this process, we construct UniMM-Chat, a high-quality multimodal instruction tuning dataset containing over 1.1M instructions. We conduct a series of experiments to demonstrate the effectiveness of our data construction pipeline and the resultant UniMM-Chat dataset. Besides, we construct the UniMM-Bench benchmark to evaluate MLLMs' abilities in reasoning and world knowledge. Specifically we collect questions from existing VL benchmarks and leverage GPT-4 to score the model output.


In general, we summarize our contribution as follows:
\begin{itemize}
    \item We propose a novel architecture, Muffin, which reformulates pre-trained VLMs as bridges between vision modules and LLMs. Muffin achieves state-of-the-art performance among existing baselines on a wide range of tasks.
    \item We construct a knowledge-intensive multimodal instruction tuning dataset, UniMM-Chat, which is constructed by requiring ChatGPT to generate dialogues given merged information from different datasets.
    \item We construct the UniMM-Bench benchmark to evaluate the overall capability of MLLMs involving diverse tasks and evaluate Muffin and other MLLM models on it.
    \item We open-source Muffin, UniMM-Chat, and UniMM-Bench to the community.
\end{itemize}

\begin{table}
\centering
\scalebox{0.9}{
\begin{tabular}{l|ccc}
\toprule
Model  & \makecell[c]{Visual \\ Encoder} & \makecell[c]{Bridge \\ Module}  &  LLM  \\ \midrule
BLIP-2        &  ViT            &  Q-Former       &   Flan-T5   \\ 
MiniGPT-4  &  ViT  & Q-Former &  Vicuna-13B \\ 
VisualGLM     &  ViT            &  Q-Former       &  ChatGLM    \\ 
Ziya-Visual &  ViT            &  Q-Former       &  Ziya-LLaMA-13B\\ 
mPLUG-owl &  ViT            &  Q-Former &  LLaMA-7B  \\ 
InstructBLIP &  ViT            &   Q-Former   &   Vicuna-13B    \\ 
LLAVA  &     ViT     &       Linear       &     Vicuna-13B    \\ 
Muffin   &    \multicolumn{2}{c}{BEiT-3}  &     Vicuna-13B    \\ 
\bottomrule
\end{tabular}}
\caption{Summary of the structure of existing MLLMs.}
\label{tab:structure_baseline}
\end{table}

\section{Related Work}
\smallskip
\textbf{Vision Language Models}
The research of pretrained VLMs has been a hot topic for years. These models are pre-trained on a large scale of image-text pairs to achieve the alignment between visual and text modalities. Some work focuses on improving the training objectives, such as contrastive loss \citep{radford2021learning, jia2021scaling}, masked data modeling \citep{wang2023image}, and image-text matching \citep{li2021align}. Some work devotes to optimizing the model architecture, such as UNITER \citep{chen2020uniter} and VinVL \citep{zhang2021vinvl}, and the recent unified transformer architecture VLMo \citep{bao2022vlmo}. Based on these techniques, several large-scale VLMs are proposed, such as Florence \cite{yuan2021florence} and BEiT-3 \citep{wang2023image}. These models exhibit good performance on VL tasks while lacking capabilities in following human instructions.

\smallskip
\textbf{Multimodal Large Language Models}
MLLMs aim to bridge a visual module with pre-trained LLMs for multimodal interaction. The pioneering work, BLIP-2 \citep{li2023blip}, introduces the Q-Former architecture, a shallow transformer to align the visual feature from the frozen visual encoder with LLMs. Subsequent works largely adopt the Q-Former architecture, such as MiniGPT-4 \citep{zhu2023minigpt}, VisualGLM \citep{du2022glm}, Ziya-Visual \citep{fengshenbang}.
LLaVA \citep{liu2023visual} employs a linear layer to map the visual feature from the frozen vision encoder into the embedding space of pre-trained LLM \citep{vicuna2023}. mPLUG-Owl \citep{ye2023mplugowl} leverages a modified Q-Former module to align the vision encoder CLIP with LLM using both text-only and multimodal instruction tuning datasets. InstructBLIP \citep{instructblip} improves the Q-Former, obtaining the instruction-aware visual features by inputting the instruction into the Q-Former as well. Table \ref{tab:structure_baseline} summarizes the detailed structure of these models.

\smallskip
\textbf{Multimodal Instruction Tuning Datasets}
To equip MLLMs with potent instruction-following capabilities, several multimodal instruction tuning datasets are proposed. MiniGPT-4 \citep{zhu2023minigpt} proposes using ChatGPT to rewrite the image description and collect nearly 3.5K instruction instances. InstructBLIP \citep{instructblip} formulate 26 publicly available datasets of different tasks with handcrafted templates for each dataset. LLaVA \citep{liu2023visual} proposes to leverage GPT-4 to write instructions for three different categories, including detail description, complex reasoning, and conversation, given annotations of images from the COCO dataset \citep{lin2014microsoft}. Though the generated data of LLaVA is more diverse compared with MiniGPT-4, the instructions are still scarce in knowledge since annotations from only one dataset can hardly give a comprehensive understanding of images.

\section{Muffin Framework}

\subsection{Architecture}
\begin{figure}
    \centering
    \includegraphics[width=0.75\columnwidth]{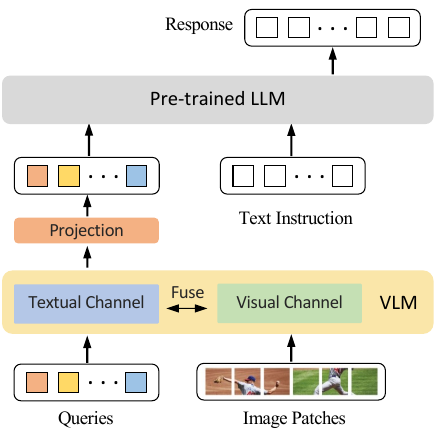}
    \caption{Architecture of our proposed Muffin. A sequence of trainable query vectors and the image patches are input into the textual and visual channels of VLM, respectively. By deeply fusing in the VLM blocks, the output from the textual channel serves as the summarized image feature.}
    \label{fig:architecture}
\end{figure}

The architecture of our proposed Muffin is shown in Figure ~\ref{fig:architecture}. Instead of training a separate module to connect the vision encoder and LLMs, Muffin directly utilizes a pre-trained VLM model, denoted as $G$, to summarize the visual representation for LLMs, denoted as $F$. Commonly, VLMs consist of a visual channel and a text channel, which are deeply fused with each other to achieve modality alignment. By extensive pre-training in large-scale V-L datasets, VLMs inherently excel in serving as the ``out-of-the-box'' bridge for LLM. In this work, we leverage BEiT-3 \citep{wang2023image} as VLM backbone, which is pre-trained with masked data modeling and achieves good performance on many vision and vision-language tasks. 

To leverage VLM for visual features extraction, Muffin introduces a sequence of trainable query vectors in the text embedding space of VLM, denoted as $\boldsymbol{Q} = [\boldsymbol{q}_1, \boldsymbol{q}_2, \cdots, \boldsymbol{q}_n], \boldsymbol{q}_n \in \mathbb{R}^d$, where $n$ is the number of trainable query vectors, $d$ is the hidden size of VLM $G$. These query vectors $\boldsymbol{Q}$ and the image $X_v$ are input into the text and vision channels, respectively. Within each block of the Transformer, to deeply fuse two modalities, the hidden states from each channel will perform both self-attention and cross-attention with each other.

After deep fusion between the trainable query vectors $\boldsymbol{Q}$ and the image, the final output in the last layer corresponding to the query vectors' position effectively captures a visual feature of the input image. 
This progression can be succinctly expressed as an end-to-end formulation:
\begin{equation}
\label{eq:q_l}
    Z_{v} = G(X_v, \boldsymbol{Q}_{\vartheta}).
\end{equation}




\begin{figure*}[t]
    \centering
    \includegraphics[width=0.97\linewidth]{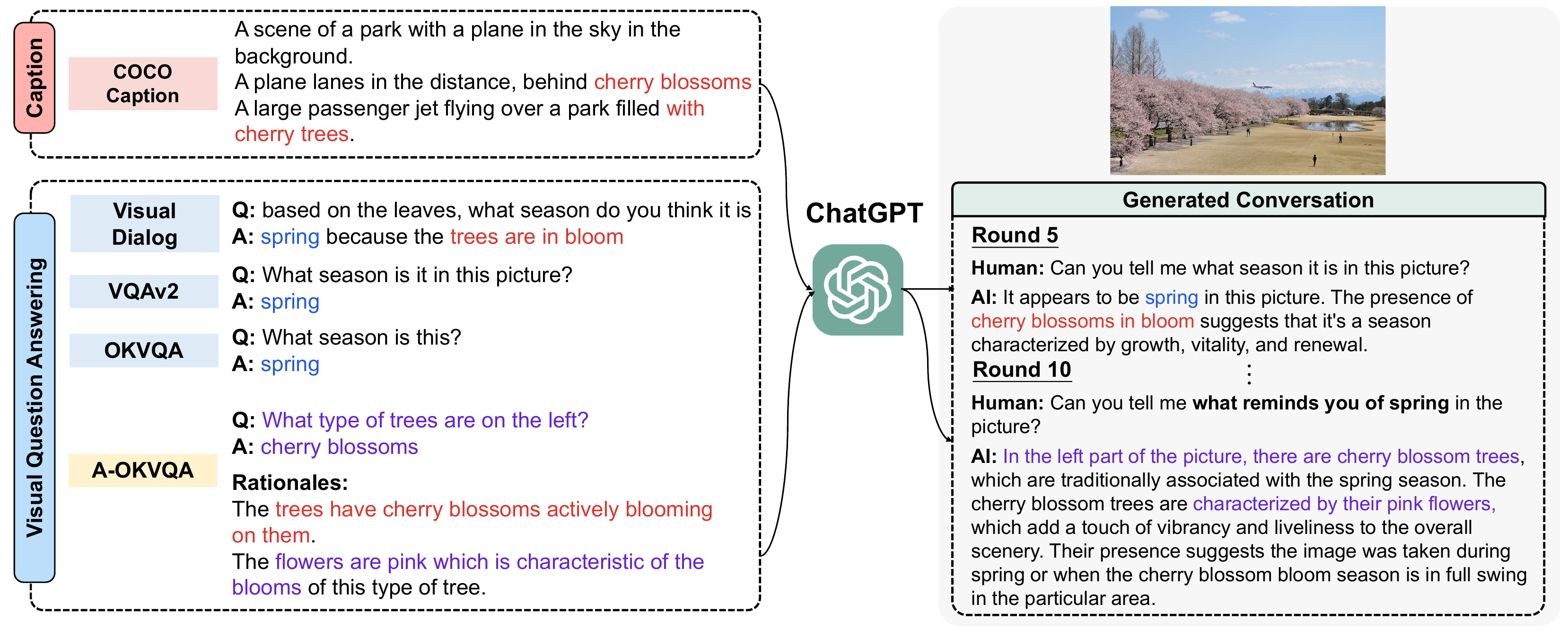}
    \caption{Demonstration of the framework designed for constructing the UniMM-Chat dataset. UniMM-Chat combines various VL datasets to generate knowledge-intensive dialogues. Text spans highlighted in colors indicate different knowledge from origin annotations which are required to answer the questions.} 
    \label{fig:cocosft_construction_framework}
\end{figure*}

Subsequently, we apply a fully connected projection layer to transform the perceived visual feature $Z_v$ into the embedding space of the pre-trained LLM $H_v =\boldsymbol{W}_\xi\cdot Z_v$, which will serve as the prefix context for the LLM and concatenate with the text embedding of $X_t$ as the final input to be forwarded to the LLM.

\subsection{Pre-training}
\label{sec:pretrain}
Following the most existing MLLMs \cite{liu2023visual, instructblip}, we first conduct pre-training on an extensive number of image-text pairs to align the VLM $G$ and the LLM $F$. Since the text data used in this stage share relatively simple formats, we freeze the parameters of LLM during pre-training to retain the powerful knowledge and complex reasoning ability of LLM. For a pair of images and text $(X_v, X_t)$, we randomly select an instruction $X_{\text{ins}}$ from a pre-defined set, as used in \citet{liu2023visual}. This instruction, such as ``Describe the image briefly.'', serves to prefix the caption and narrow the gap in training data format between the current stage and following instruction tuning. The training objective is to maximize the probability of target response given the instruction $X_{\text{ins}}$ and input image $X_v$, shown as:
\begin{equation} 
\label{eq:lm_loss}
    \mathcal{L} = \sum\limits_{i=1}^{k} \log p(x_i|\boldsymbol{H}_v, X_{\text{ins}}, X_{t, <i})
\end{equation}


\subsection{Multimodal Instruction Tuning}
\label{sec:sft}
While naive multimodal pre-trained models exhibit the capacity to comprehend the content of input images and generate concise captions, they often lack the ability to execute intricate tasks based on human instructions. As a result, we proceed to undertake further multimodal instruction tuning.

Unlike the previous stage, we make the LLM trainable during the instruction tuning process to harness the full potential of high-quality instructional data. We structure each data instance in the form of a conversational snippet following Vicuna \citep{vicuna2023} and train the model to decode tokens of the answer spans. We utilize the same training objective, as represented by Equation \eqref{eq:lm_loss}, used in the pre-training stage.


\section{UniMM-Chat}
To construct high-quality and diverse instruction tuning datasets with comprehensive image descriptions, we propose the UniMM-Chat dataset, which consists of 1.1M diverse instructions. We incorporate complementary annotations from different VL datasets and employ ChatGPT to generate multi-turn dialogues corresponding to each image. As shown in Figure \ref{fig:cocosft_construction_framework}, the incorporated annotations furnish a richer image context and effectively empower ChatGPT to generate more knowledge-intensive conversational datasets.

\subsection{Dataset Construction}
\label{sec:data-construction}
Five commonly utilized VL datasets, as outlined in Table \ref{tab:cocosft_construction_input_statistic}, serve as seeds to craft multimodal instructions. As images in these five datasets are drawn from COCO \citep{lin2014microsoft}, we first aggregate the annotations for each image from the seed VL datasets. For VQAv2 \citep{balanced_vqa_v2}, OKVQA \citep{okvqa}, and Visual Dialog \citep{visdial}, we use the annotation of both questions and their corresponding answers. For AOKVQA \citep{AOKVQA}, we use the question-answer pair and the annotated rationales. Five captions from COCO \citep{lin2014microsoft} are directly employed as fundamental descriptions for each image. 

Next, these annotations are meticulously structured into a refined format, incorporating some additional human-written few-shot learning instances. These elements are collectively presented as prompts, prompting ChatGPT to generate multi-turn dialogues centered on the respective images. We refer readers to the Appendix for the prompts we used during data construction.


\begin{figure}[t]
    \centering
    \includegraphics[width=0.75\columnwidth]{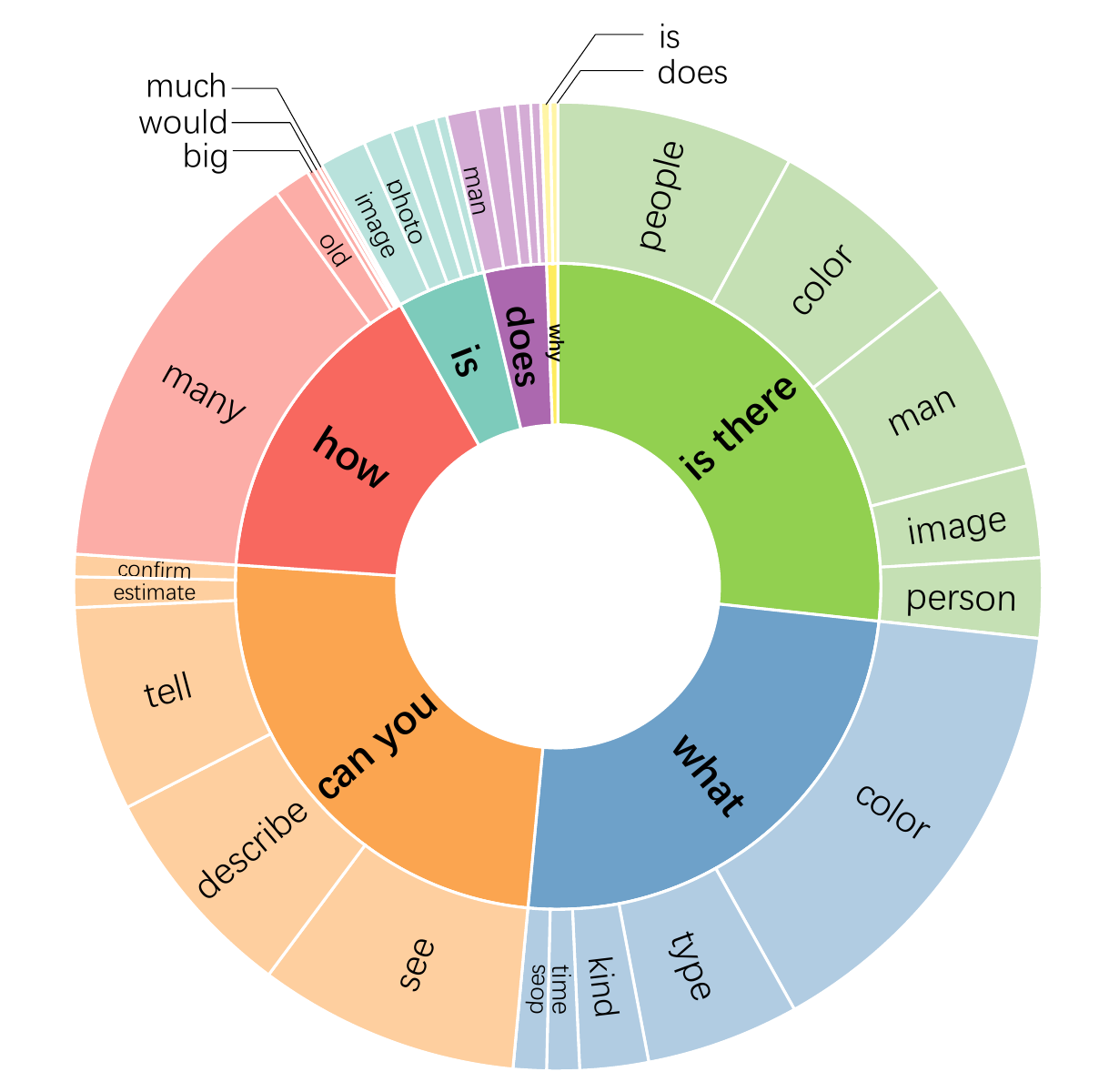}
    \caption{Instruction distribution in UniMM-Chat.}
    \label{fig:cocosft_question_distribution}
\end{figure}

\subsection{Dataset Statistics}

In total, we collect 117,238 dialogues, with an average of 9.89 turns per dialogue. Each dialogue is associated with one distinct image. To quantify the dataset's diversity, we follow \citep{wang-etal-2023-self-instruct} and parse the question types and their direct nouns or verb with Berkeley Neural Parser tool \citep{stern2017minimal}. We plot the seven most common question types and their top direct noun objects or verbs, which account for 44\% instructions in UniMM-Chat in Figure \ref{fig:cocosft_question_distribution}.  This plot underscores the considerable breadth of intents and formats in UniMM-Chat.

\subsection{UniMM-Bench}
We propose UniMM-Bench, a question-answering benchmark designed for MLLMs to evaluates the abilities involving reasoning and world knowledge. As traditional exact-matched accuracy is not suitable for evaluating MLLMs, which often respond a complete sentence to answer questions, we leverage GPT-4 to score the generated answer. Considering the evaluation cost, we sample one hundred samples from the test set of OKVQA \citep{okvqa}, AOKVQA \citep{AOKVQA}, GQA \citep{hudson2019gqa} and VQAv2 \citep{balanced_vqa_v2}, respectively, whose annotations has undergone meticulous inspection. This benchmark evaluates the capabilities of MLLMs in reasoning, commonsense, and world knowledge \citep{AOKVQA}.

\begin{table}[t]
    \centering
    \scalebox{0.9}{
    \begin{tabular}{lrr}
    \toprule
       VL dataset  & \#Images & \#Annotation  \\
       \midrule
        VQAv2 \cite{balanced_vqa_v2}   & 123,287 & 658,111\\
        OKVQA \citep{okvqa}  & 14,031 & 14,055 \\
        AOKVQA \citep{AOKVQA}  & 17,662 & 18,201 \\
        Visual Dialogue \citep{visdial}  & 123,287 & 1,232,870 \\
        COCO Caption \citep{lin2014microsoft}  & 123,287 & 616,767 \\
\bottomrule    
    \end{tabular}}
    \caption{Statistics of source vision-language datasets to construct UniMM-Chat.}
    \label{tab:cocosft_construction_input_statistic}
\end{table}


\begin{table*}[t]
    \centering
    \resizebox{\linewidth}{!}{
    \begin{tabular}{l|ccccc cccc c}

\toprule

\multirow{2}{*}{\textbf{Model}} & \multicolumn{5}{c}{\textbf{UniMM-Bench}} & \multicolumn{4}{c}{\textbf{LLaVA Test Set}} & \multirow{2}{*}{\textbf{ALL}}\\

\cmidrule(lr){2-6} \cmidrule(lr){7-10}   & OKVQA & 
  AOKVQA & GQA & VQAv2 & {\bf AVG} & Con & CR & DD & {\bf AVG} & \\

\midrule

VisualGLM \citep{du2022glm} &  56.2 & 56.6 & 59.2 & 63.9 & 59.0 & 65.8 & 80.6 & 64.5 & 70.3 & 60.6\\
MiniGPT-4 \citep{zhu2023minigpt}  & 59.2 & 56.0 & 66.0 & 58.9 & 60.0 & 65.3 & 75.6 & 66.3 & 69.1 & 61.8 \\
mPLUG-owl \citep{ye2023mplugowl} & 63.9 & {63.3} & 63.3 & 64.1 & 63.6 & 69.0 & 84.1 & 59.0 & 70.8 & 65.1 \\
Ziya-Visual \citep{fengshenbang} &  67.3 & 65.6 & {68.2} & 67.0 & 67.0 & 82.3 & {90.2} & 71.2 & 81.3 & 69.9\\
LLaVA \citep{liu2023visual} & {68.8} & 64.2 & 68.1 & 67.7 & 67.2 & \underline{83.0} & \textbf{96.5} & {75.0} & 84.9 & 70.7 \\
InstructBLIP \citep{instructblip} &  67.0 & 64.9 & 67.6 & \underline{75.5} & \underline{68.8} & 82.2 & 90.2 & 68.4 & 80.7  & 71.1 \\

\midrule
Muffin  (w/o UniMM-Chat) & \underline{69.1} & \underline{68.4} & \underline{70.1} & 66.8 & 68.6 & 82.2 & \underline{96.0} & \textbf{77.5}  & \underline{85.3} & \underline{71.9}\\
Muffin & {\bf 72.8} & {\bf 69.8} & {\bf 72.9} & {\bf 77.9} & {\bf 73.4} & {\bf 83.5}  & \textbf{96.5} & \underline{77.2} & \textbf{85.7} & \textbf{75.8} \\
 \bottomrule
    \end{tabular}
    }
    \caption{Performance of our proposed Muffin and baselines on UniMM-Bench and LLaVA Test Set. AVG: average of scores. Con: conversation category. CR: complex reasoning category. DD: detail description category. We report the average performance of three trials for each model to improve the stability.}
    \label{tab:main_results}
\end{table*}

\section{Experiments}
\subsection{Experimental Settings}
\smallskip
\textbf{Evaluation Details}. We evaluate MLLMs on our proposed UniMM-Bench and LLaVA test set\citep{liu2023visual}. The LLaVA test set consists of 90 questions from three categories spanning conversation, complex reasoning, and detail description. UniMM-Bench mainly evaluates the model abilities in reasoning and world knowledge, while the LLaVA test set evaluates the model performance on multimodal conversational.
We leverage GPT-4 to score the model output based on the ground truth answers. We empirically verified the scores of GPT-4 are well aligned with human judgment. We refer readers to the Appendix for complete prompts.

\smallskip
\noindent\textbf{Training Details}.
The pre-training of Muffin is performed with 180M image-text pairs collected from Visual Genome \citep{krishna2017visual}, COCO \citep{lin2014microsoft}, CC3M \citep{sharma2018conceptual}, CC12M \citep{changpinyo2021cc12m} and LAION-COCO \citep{laioncoco} and lasts for 100K steps with batch size of 2048 and learning rate of 1e-4. For instruction tuning, we use both the LLaVA-Instruct-150K and UniMM-Chat instruction tuning dataset. The training lasts for 3200 steps with batch size of 512 and learning rate of 2e-5. We adopt the resolution of 448 during pre-training and 672 during the instruction tuning stage.

\smallskip
\noindent\textbf{Baselines}. We compare our method with a series of existing strong baselines:

\begin{itemize}
\item  \textbf{MiniGPT-4}: MiniGPT-4 \citep{zhu2023minigpt} is one of the earliest open-source trials of MLLMs, which is fine-tuned on over 3.5K simple instructions that require model to generate image descriptions.
\item  \textbf{VisualGLM}: VisualGLM \citep{du2022glm} is a bilingual multimodal assistant model built upon ChatGLM-6B and the vision encoder of BLIP2 which devises complex feature alignment training process.
\item  \textbf{Ziya-Visual}: \citep{fengshenbang} is a bilingual multimodal assistant model based on Ziya-LLaMA-13B and pre-trained visual encoder of BLIP-2. 
\item  \textbf{mPLUG-owl}: mPLUG-owl \citep{ye2023mplugowl} is a multimodal assistant model based on CLIP ViT-L/14 and LLaMA-7B, which use both text-only and multimodal instruction tuning datasets. 
\item \textbf{InstructBLIP}: InstructBLIP \citep{instructblip} constructs a multimodal instruction tuning dataset based on 26 public datasets by apply pre-defined templates to directly formulate these datasets into a unified format. They devise a novel instruction-aware Q-Former and train the model on the proposed dataset.
\item  \textbf{LLaVA}: LLaVA \citep{liu2023visual} constructs 150K multimodal instructions based on the COCO dataset. It simply leverages a linear projector to connect the vision encoder and LLM. 
\end{itemize}


\subsection{Main Results}

Table \ref{tab:main_results} presents the performance of Muffin and baselines on UniMM-Bench and LLaVA test set. On both of these two benchmarks, Muffin achieves the state-of-the-art performance and significantly surpasses all baselines. 

Based on these experimental results, we have the following observations:
\begin{itemize}
    \item Compared with LLaVA, Muffin achieves an impressive 5.1-point advancement on average. Also, even when employing the same LLaVA-Instruct-150K dataset for instruction tuning, akin to LLaVA's training, Muffin still achieves better results with 1.2-point, demonstrating the effectiveness of the Muffin framework.
    \item Compared with InstructBLIP, Muffin exhibits substantial performance enhancements over InstructBLIP.  Specifically, despite directly training on OKVQA, AOKVQA, and VQAv2, InstructBLIP achieves lower performances on these datasets compared with Muffin, especially on OKVQA and AOKVQA, which contains limited annotations. This indicates simply combining training samples of different datasets is sub-optimal for the model to learn a wide range of  knowledge. On the LLaVA test set, we hypothesize the limited format of responses in the training data harms the generation ability and consequently results in InstructBLIP significantly lag behind Muffin.
    \item Excluding UniMM-Chat from the training set leads to a substantial performance drop across all visual question-answering tasks. This emphasizes the pivotal role of UniMM-Chat in equipping MLLMs with the skills to effectively address a variety of task types.
\end{itemize}
These results collectively demonstrate the effectiveness of the Muffin framework and highlight the crucial role of UniMM-Chat. Enriched by UniMM-Chat, Muffin exhibits strong reasoning capabilities and abundant knowledge.
\begin{figure}
    \centering
    \includegraphics[width=0.87\columnwidth]{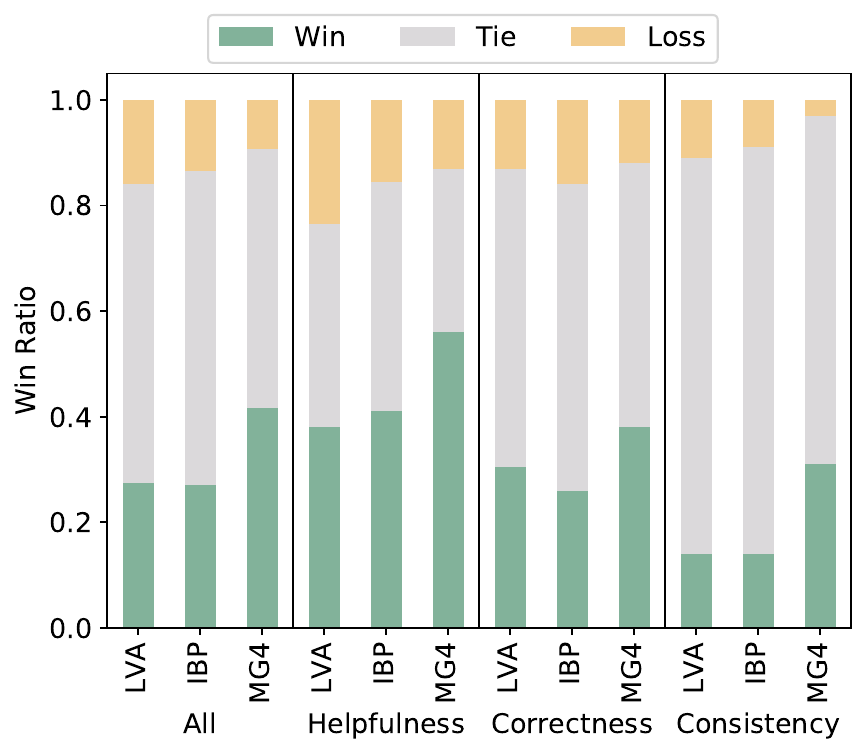}
    \caption{Human evaluation winning ratio of Muffin compared with different baseline models. LVA: LLaVA. IBP: InstructBLIP. MG4: MiniGPT-4.}
    \label{fig:human_eval}
\end{figure}

\begin{figure*}
    \centering
    \includegraphics[width=0.95\linewidth]{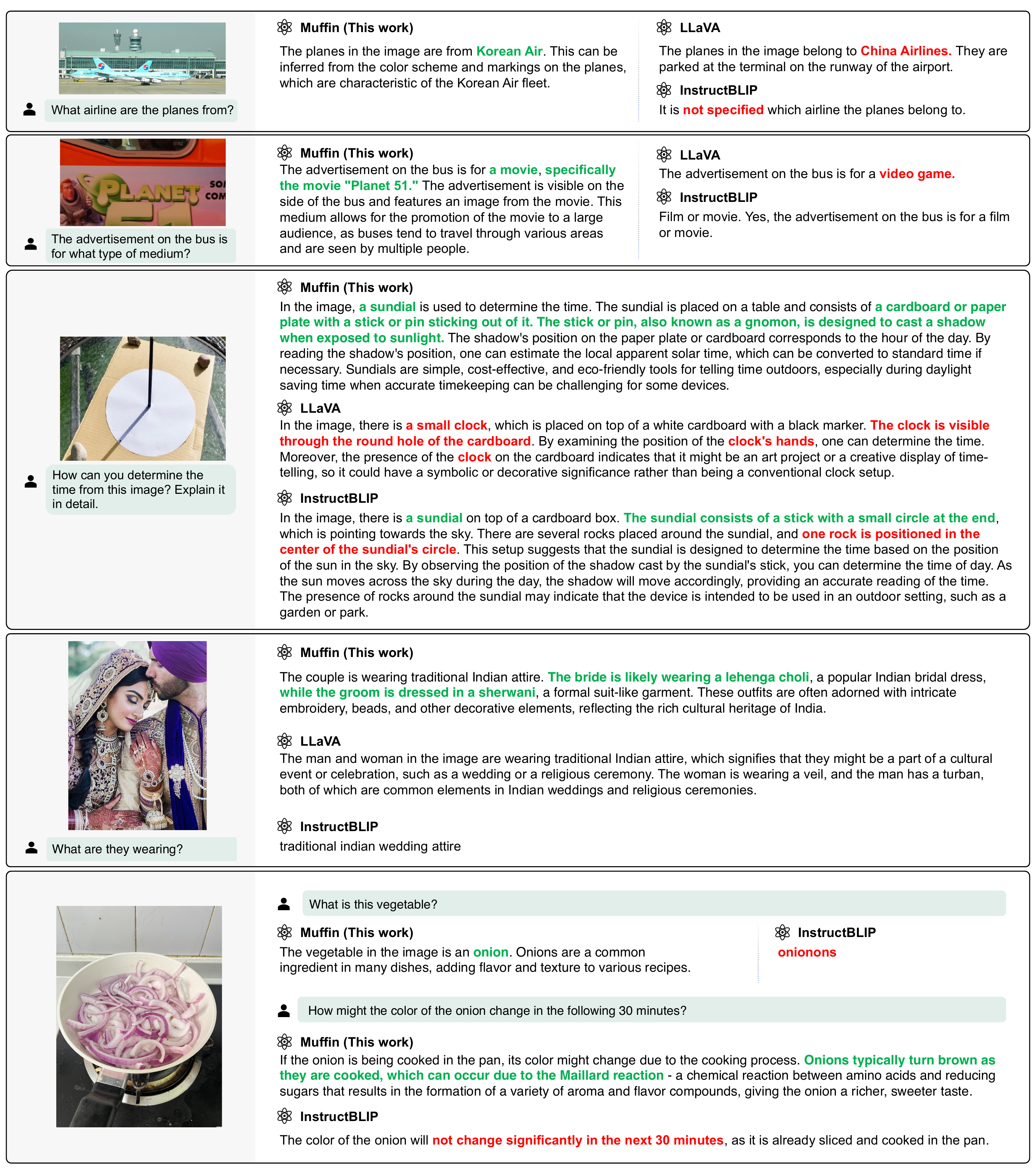}
    \caption{Examples generated by Muffin and other baselines.}
    \label{fig:show_case}
\end{figure*}

\subsection{Human Evaluation}

For a more comprehensive analysis, we conduct a pair-wise human evaluation of different models on a diverse range of instructions. Specifically, we randomly sample eighty samples from UniMM-Bench and twenty samples from the LLaVA test set. We recruit six well-educated annotators and present them with the answer pairs generated by Muffin and other three baselines, LLaVA, InstructBLIP, and MiniGPT-4. We assure the model names are hidden during the whole evaluation process.   Annotators are required to decide which answer is better in each pair, based on three criteria: Helpfulness, Correctness, and question-answer Consistency. More details are introduced in the Appendix. The evaluation results are shown in Figure \ref{fig:human_eval}. Muffin outperforms all other models on all metrics. The evident advantage of Helpfulness and Correctness originates from training on the knowledge-intensive dialogues from the UniMM-Chat dataset. Specifically, we find the MiniGPT-4 \citep{zhu2023minigpt}, which is trained with only a few thousand simple instruction samples, usually responds that is not related to the question and consequently obtains the lowest consistency win ratio. We refer readers to the Appendix for the detail statistics of human evaluation results comparing Muffin with baseline models.

\subsection{Ablation Results}



\begin{table}[]
    \centering
    \resizebox{\linewidth}{!}{
    
    \begin{tabular}{lcc}
       \toprule
       Instruction Data & UniMM-Bench & LLaVA Test Set\\
       \midrule
        Origin Datasets & 72.5 & 36.6\\
        \midrule
        UniMM-Chat-sep & 70.7 & 65.6\\
        UniMM-Chat  & \textbf{72.6} & \textbf{78.6} \\
        \bottomrule
    \end{tabular}
    }
    \caption{Muffin trained with different instruction tuning dataset settings. UniMM-Chat-sep is a variant of UniMM-Chat that is constructed without merging annotations from different VL datasets.}
    \label{tab:sft_ablation}
\end{table}

\begin{table}[]
    \centering
    \resizebox{\linewidth}{!}{
    
    \begin{tabular}{lcc}
       \toprule
       Model & UniMM-Bench & LLaVA Test Set\\
       \midrule
        Muffin & \textbf{73.4} &  \textbf{85.7} \\
        \midrule
        \hspace{2mm} - pre-training resolution & 70.1 & 85.5 \\
        \hspace{2mm} - instruction tuning resolution  & 72.1 & 85.3 \\
        \hspace{2mm} - tune LLM & 64.0 & 68.4 \\
        \bottomrule
    \end{tabular}
    }
    \caption{Ablation results for different training settings.}
    \label{tab:training_ablation}
\end{table}

\noindent\textbf{Reformulating VL datasets}. To verify the effectiveness of the dataset construction framework, we build a variant of UniMM-Chat without merging annotations across different datasets, named UniMM-Chat-sep. More details of UniMM-Chat-sep are presented in Appendix. We train multiple models using different data configurations and present the result in Table \ref{tab:sft_ablation}. On the LLaVA test set, using original datasets fails to establish a chat model due to the short text in these datasets. Incorporating UniMM-Chat-sep, while resulting in improved performance compared to using the original datasets, still yields suboptimal results owing to the limited information available during construction. On UniMM-Bench, directly using the original datasets corresponds to in-domain fine-tuning, serving as the performance upper bound for the constructed dataset. The experimental results show that using UniMM-Chat does not lead to the in-domain performance drop, while using UniMM-Chat-sep will undergo a performance decline. These results emphasize the necessity of our dataset construction framework.



\smallskip
\noindent\textbf{Training Settings.} We also analyze the effect of some training settings and present the results in Table \ref{tab:training_ablation}. As for the image resolution, when decreasing the input image resolution during the pre-training (224) and instruction tuning stage (448), the performance has a distinct drop on UniMM-Bench, since detailed image information necessary to solve complex multimodal tasks is hard to be retrained with low resolution. Besides, we also observed freezing the LLM during the instruction tuning process can limit the model's abilities on all evaluated datasets.




\subsection{Qualitative Results}

Benefiting from the deep fusion within the VLM and knowledge intensive instruction data from UniMM-Chat, Muffin can effectively activate the knowledge embedded in the LLM and generate more helpful response to open-ended questions. Figure \ref{fig:show_case} shows some examples of responses from Muffin, LLaVA, and InstructBLIP for demonstration. In the first example, our model accurately identifies the country of origin of the airplane based on visual details, while both LLaVA and InstructBLIP fail to provide the correct answer, highlighting our model's superiority in comprehending and identifying image details. In the second example, our model combines textual cues from the image with its inherent knowledge to generate a more accurate response. Moreover, in the fourth example, Muffin correctly identifies the exact name of the traditional Indian attire shown in the image. Except having the ability to answer questions with a broad range of knowledge, Muffin can also generate more helpful responses. In the last example show in \ref{fig:show_case}, though both pointing the vegetable is onion, Muffin gives more detail and helpful response. 

\section{Conclusion}
In this paper, we present Muffin, an innovative framework to directly utilize pre-trained VLMs to bridge visual signals and LLMs. Also, we develop a new paradigm to build a multimodal instruction tuning dataset by merging annotations from different datasets describing the same images. In this way, we construct a high-quality and diverse multimodal instruction tuning dataset, UniMM-Chat. We perform comprehensive experiments to demonstrate the effectiveness of Muffin and UniMM-Chat, which shows that Muffin achieves state-of-the-art performance on a wide range of tasks. In the future, we will apply Muffin framework and UniMM-Chat dataset to more combinations of VLMs and LLMs.



\bibliography{aaai24}

\appendix

\section{Prompts}

In this section, we list details of all prompts we use in this work for reproducibility, including the prompts to construct UniMM-Chat, the prompt to evaluate for UniMM-Bench and the prompt used to pre-train Muffin.

\subsection{UniMM-Chat Construction Prompts}

\begin{figure}[t]
    \centering
    \includegraphics[width=0.75\columnwidth]{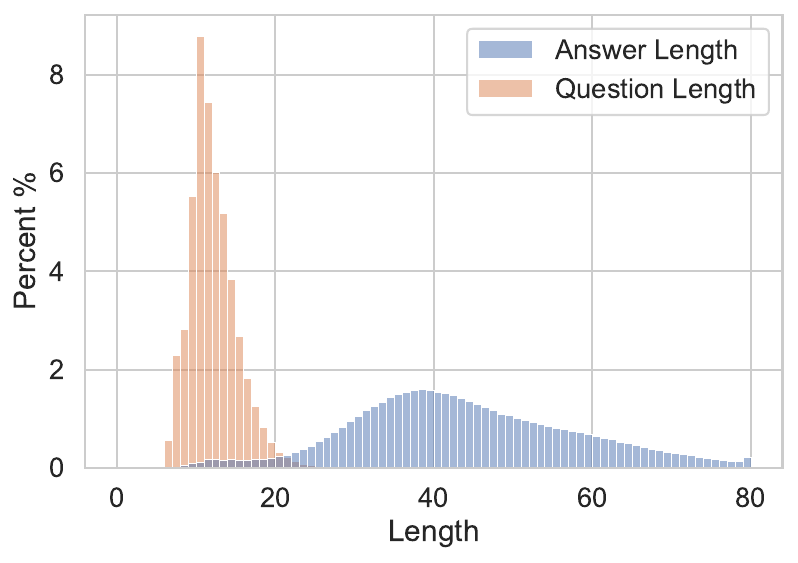}
    \caption{Length distribution of the generated questions and answers in UniMM-Chat.}
    \label{fig:cocosft_length_distribution}
\end{figure}

We show the full prompt we used to require ChatGPT to generate high quality knowledge-intensive dialogues for UniMM-Chat in Table \ref{tab:cocosft_construction_prompt}. We present the raw prompt and how we organize few human annotated demonstrations together with the raw prompt. We also list the prompt to amass origin annotations from different VL datasets in Table \ref{tab:cocosft_construction_input_format}, which generate the \textit{input} to be used in Table \ref{tab:cocosft_construction_prompt}. We also present the question and answer length distribution of UniMM-Chat in Figure \ref{fig:cocosft_length_distribution} for reference.

\begin{table}[h!]
\centering

\begin{minipage}{0.99\columnwidth}\vspace{0mm}    \centering

\begin{tcolorbox} 
    \small
    \hspace{-6mm}

[Image statements]

\{VQAv2\_qas\}

\{OKVQA\_qas\}

\{AOKVQA\_qas\}

\{VisualDialog\_qas\}

\vspace{3mm}

[Image information]

\{AOKVQA\_raionales\}

\vspace{3mm}

[ Image description ]

\{COCO\_captions\}

\vspace{3mm}

[Conversation]

\end{tcolorbox}
    
\vspace{-2mm}
\caption{The template to amass annotations from different VL datasets.}
\label{tab:cocosft_construction_input_format}
\end{minipage}
\end{table}

\begin{table*}[h!]
\centering

\begin{minipage}{0.99\linewidth}\vspace{0mm}    \centering

\begin{tcolorbox} 
    \small
    \hspace{-6mm}

Prompt:

messages = [ {``role'': ``system'', ``content'': f``You are an AI visual assistant, and you are seeing a single image. What you see are provided with [Image statements], as well as the [Image information] and [Image description] in several sentences, describing the same image you are looking at. You should pretend not seeing [Image statements], [Image information], [Image description], etc. Instead, ask and answer all questions as you are only seeing the image.

\hspace{3mm}

Design a 10 rounds conversation between you and a person asking about this photo. The answers should be in a tone that a visual AI assistant is seeing the image and answering the question.
Ask diverse questions and give corresponding answers.

\hspace{3mm}

Include questions asking about the visual content of the image, including the object types, object actions, object locations, relative positions between objects, etc. Only include questions that have definite answers:

(1) one can see the content in the image that the question asks about and can answer confidently.

(2) one can determine confidently from the image that it is not in the image.

\hspace{3mm}

Do not ask any question that cannot be answered confidently. Do not add unprovided details to answer the questions.

Also include complex questions that are relevant to the content in the image, for example, asking about background knowledge of the objects in the image, asking to discuss about events happening in the image, etc. Again, do not ask about uncertain details.

You should do your best to fully cover the image content in the conversation. Try to ask questions using special interrogative sentences.

Provide detailed answers when answering questions if necessary. For example, give detailed examples or reasoning steps to make the content more convincing and well-organized.  You can include multiple paragraphs if necessary. Double-check to ensure your answer is correct and consistent with the image and the previous conversation. Stay within the scope of the image, refraining from introducing any information not present in the visual content. Again, do not add unsupported details to answer the questions. Answer all questions as you are only seeing the image.''} ] 

\vspace{3mm}

\texttt{for sample in fewshot\_samples: }

\hspace{4mm}\texttt{messages.append(\{`role':`user', `content':sample[`input']\})} 	

\hspace{4mm}\texttt{messages.append(\{`role':`assistant', `content':sample[`output']\})}

\hspace{4mm}\texttt{messages.append(\{`role':`user', `content':query\})}
\end{tcolorbox}
    
\vspace{-2mm}
\caption{The detail prompt used to 
 guide ChatGPT to generate conversations.}
\label{tab:cocosft_construction_prompt}
\end{minipage}

\end{table*}

\subsection{UniMM-Bench Evaluation Prompt}

We list the full prompt we used to evaluate the performance of models on UniMM-Bench in Table \ref{tab:evaluation_prompt}. To enable GPT-4 generate more accurate scores, we put both the ground truth answer and other related annotations into the prompt. Specifically, we list all human answers and corresponding confidences from VQAv2 \citep{balanced_vqa_v2} and rationales from AOKVQA \citep{AOKVQA} into the prompt.

\begin{table}[h!]
\centering

\begin{minipage}{0.99\columnwidth}\vspace{0mm}    \centering

\begin{tcolorbox} 
    \small
    \hspace{-6mm}

[Question]

\textit{\{question\}}

\vspace{3mm}
[Assistant Response]

\textit{\{response\}}

[End of Assistant Response]

\vspace{3mm}

[System]

We would like to request your feedback to evaluate the performance of an AI assistant in the response to an user question displayed above. The AI assistant is asked to look the image and answer the question. You need to give an overall score to the assistant's response to the question on a scale of 1 to 5, where a higher score indicates better overall performance. Please first output a single line containing only one value indicating the score for the assistant.

\vspace{1mm}

In the subsequent line, please provide a comprehensive explanation of your evaluation. 

\vspace{1mm}

We will give you some additional information about the image and question for reference in the following (such as the expected answer, human answers and hints given by annotators). Note that the assistant can only see the image content and question text, all other reference information are used to help you better understand the question and content of the image only. The major criteria is the correctness of the answer, you don't have to care about the conciseness or structure or other irrelevant factors of the answer.

\vspace{3mm}

[Expected Answer]

\textit{\{ground truth answer\}}

\vspace{3mm}

[Human Answers]

\textit{\{human answers and rationales from origin datasets\}}

\end{tcolorbox}
    
\vspace{-2mm}
\caption{GPT-4 evaluation prompt used to evaluate UniMM-Bench.}
\label{tab:evaluation_prompt}
\end{minipage}

\end{table}

\begin{table}[]
    \centering
    \resizebox{\columnwidth}{!}{
    \begin{tabular}{lc}
\toprule
\textbf{Caption-generation Instruction} \\
\midrule
Describe the image concisely.\\
Provide a brief description of the given image. \\
Offer a succinct explanation of the picture presented.\\
Summarize the visual content of the image. \\ 
Give a short and clear explanation of the subsequent image. \\
Share a concise interpretation of the image provided. \\
Present a compact description of the photo's key features.\\
Relay a brief, clear account of the picture shown. \\
Render a clear and concise summary of the photo. \\
Write a terse but informative summary of the picture. \\
Create a compact narrative representing the image presented. \\
\bottomrule
    \end{tabular}
    }
    \caption{Instructions used to generate captions during the multimodal pre-training stage.}
    \label{tab:cap_gen_instruction}
\end{table}

\subsection{Pre-training Prompts}

We adopt the pre-training prompts used in \citep{liu2023visual} for the training of Muffin. The full list of prompts to generate caption are listed in Table \ref{tab:cap_gen_instruction}.

\section{UniMM-Chat-sep}

We construct another version of UniMM-Chat for experimental usage, namely UniMM-Chat-sep, which is constructed without merging annotations from different VL datasets. Specifically, we reuse the same prompt (see Table \ref{tab:cocosft_construction_prompt}) of UniMM-Chat to construct dialogues for each datasets except COCO Caption by removing annotations from other datasets. As for COCO Caption, since caption text is already flexible in format, we simply adopt the instructions listed in Table \ref{tab:cap_gen_instruction} following our pre-training setting. As shown in in Table \ref{tab:sft_ablation}, we find the combination of different VL datasets can yield better performance on both visual question answering and visual chat tasks. We argue this is because simply merging different datasets during training is not enough to fully explore the complementary nature of different datasets and thus give sub-optimal results.

\section{UniMM-Bench}

In order to evaluate the overall capability of MLLMs involving diverse tasks, we construct the UniMM-Bench benchmark. We randomly add 100 samples from each of the validation sets of OKVQA \cite{okvqa}, GQA \cite{hudson2019gqa}, AOKVQA \cite{AOKVQA} into UniMM-Bench. As for VQAv2 \cite{balanced_vqa_v2}, we first randomly select two sample for each type of question to keep the diversity of our benchmark, which results in 130 samples. Then we randomly choose 100 samples out of these 130 samples. Finally we combined all these 400 samples to form the UniMM-Bench benchmark.

To prevent the evaluation results being affected by data leak, we explicitly removed images in UniMM-Bench out of the image set we used to construct UniMM-Chat.

\begin{figure*}
    \centering
    \includegraphics[width=0.95\linewidth]{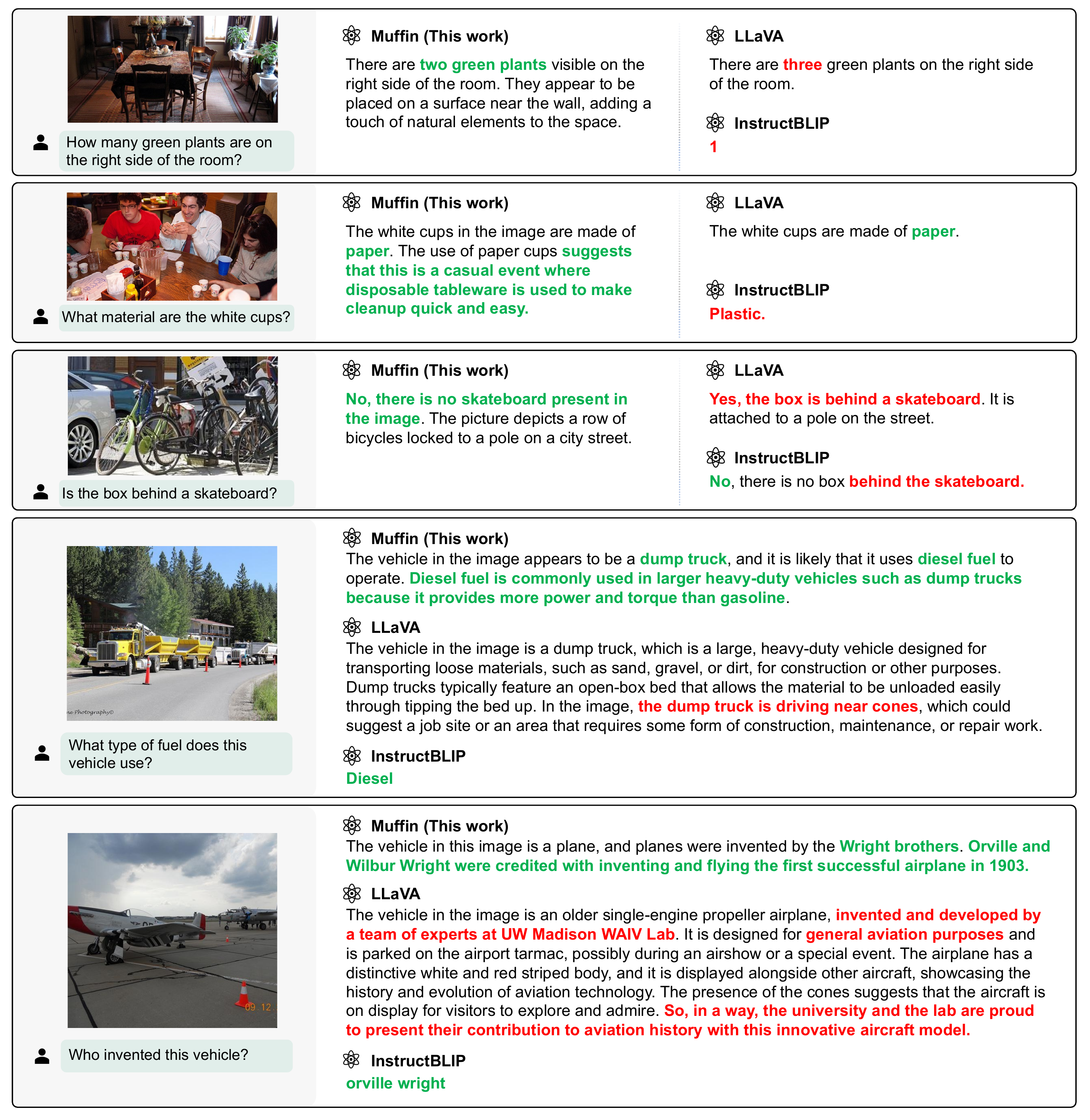}
    \caption{Examples of questions from UniMM-Bench and results generated by Muffin and other models.}
    \label{fig:appendix_case}
\end{figure*}

We list some examples from UniMM-bench and the results generated by Muffin and other baselines in Figure \ref{fig:appendix_case}. These examples show that UniMM-bench can evaluate the model abilities in both reasoning and world knowledge.

\section{Human Evaluation}

We list the detail statistics of human evaluation results in Table \ref{tab:human_eval_detail} for reference.

\begin{table}[]
    \centering
    \resizebox{\linewidth}{!}{
    
    \begin{tabular}{lrrr}
       \toprule
       Muffin vs. & Helpfulness & Correctness & Consistency\\
       \midrule
        Mini-GPT4  & \textbf{112} / 26 & \textbf{76} / 24 & \textbf{62} /\hspace{2.5mm}6 \\
        LLaVA & \textbf{76} / 47 &  \textbf{61} / 26 & \textbf{28} / 22\\
        Instruct BLIP & \textbf{82} / 31 & \textbf{52} / 32 & \textbf{28} / 18\\
        \bottomrule
    \end{tabular}
    }
    \caption{Number of samples marked as win/loss on different metrics comparing Muffin and baseline models.}
    \label{tab:human_eval_detail}
\end{table}

\end{document}